\pdfoutput=1

\documentclass[11pt]{article}

\usepackage{acl}

\usepackage{times}
\usepackage{latexsym}
\usepackage{tabularx}
\usepackage{booktabs}

\usepackage[T1]{fontenc}

\usepackage[utf8]{inputenc}

\usepackage{microtype}

\usepackage{inconsolata}

\usepackage{graphicx}

\title{L3Cube-MahaSTS: A Marathi Sentence Similarity Dataset and Models}

\author{
    Aishwarya Mirashi\textsuperscript{1,4}, Ananya Joshi\textsuperscript{3,4}, Raviraj Joshi\textsuperscript{2,4}\thanks{Correspondence: ravirajoshi@gmail.com} \\
    \textsuperscript{1}Pune Institute of Computer Technology, Pune \\
    \textsuperscript{2}Indian Institute of Technology Madras, Chennai \\
    \textsuperscript{3}MKSSS’ Cummins College of Engineering for Women, Pune \\
     \textsuperscript{4}L3Cube Labs, Pune
}

\begin{document}
\maketitle
\begin{abstract}

We present MahaSTS, a human-annotated Sentence Textual Similarity (STS) dataset for Marathi, along with MahaSBERT-STS-v2, a fine-tuned Sentence-BERT model optimized for regression-based similarity scoring. The MahaSTS dataset consists of 16,860 Marathi sentence pairs labeled with continuous similarity scores in the range of 0–5. To ensure balanced supervision, the dataset is uniformly distributed across six score-based buckets spanning the full 0–5 range, thus reducing label bias and enhancing model stability. We fine-tune the MahaSBERT model on this dataset and benchmark its performance against other alternatives like MahaBERT, MuRIL, IndicBERT, and IndicSBERT. Our experiments demonstrate that MahaSTS enables effective training for sentence similarity tasks in Marathi, highlighting the impact of human-curated annotations, targeted fine-tuning, and structured supervision in low-resource settings. 
The dataset and model are publicly shared at \url{https://github.com/l3cube-pune/MarathiNLP}.
\end{abstract}

\section{Introduction}

Semantic textual similarity (STS) \citet{cer2017semeval} refers to the task of quantifying how closely two sentences are related in meaning. Unlike surface-level methods that rely on exact word overlap, STS captures deeper semantic relationships by assessing the underlying intent or meaning, even when lexical and syntactic forms differ. This capability is essential for a wide range of natural language processing (NLP) applications such as retrieval augmented generation (RAG) \citet{lewis2020retrieval}, information retrieval \citet{karpukhin2020dense}, question answering \citet{devlin2019bert}, paraphrase detection \citet{dolan2005automatically}, and text clustering \citet{reimers2019sentence}.

While substantial progress has been made in high-resource languages like English, STS in low-resource languages remains underexplored due to the lack of annotated data \citet{agirre-etal-2012-semeval,deode2023l3cube}. For instance, Marathi, one of the most widely spoken Indian languages, suffers from a scarcity of high-quality, labeled datasets \citet{joshi2023l3cube,joshi2022l3cube}. This limits the development and evaluation of language models in real-world applications. As large language models (LLMs) continue to improve, building semantically rich resources in such languages becomes critical for advancing cross-lingual and regional NLP.

Existing STS benchmarks, such as the STS Benchmark (STSb)\footnote{https://huggingface.co/datasets/mteb/stsbenchmark-sts}, have been instrumental in evaluating sentence similarity models for English. The STSb dataset contains sentence pairs from domains like image captions, news articles, and forums, annotated with similarity scores ranging from 0 to 5. A multilingual extension, STSb Multi MT\footnote{https://huggingface.co/datasets/mteb/stsb\_multi\_mt}, provides machine-translated versions in several languages (e.g., German, Spanish, Chinese), but it lacks coverage for Indic languages such as Marathi. Furthermore, machine-translated datasets often fail to preserve cultural and linguistic nuances, resulting in reduced effectiveness for training and evaluation in low-resource settings.

To address this gap, we introduce L3Cube-MahaSTS\footnote{\url{https://huggingface.co/datasets/l3cube-pune/MahaSTS}}, a human-annotated Marathi sentence similarity dataset comprising 16,860 sentence pairs. Each sentence pair is assigned a similarity score in the range of 0 to 5. To ensure balanced label representation, the scores are grouped into six uniformly distributed buckets, with 2,810 sentence pairs per bucket. This uniform bucketing minimizes label bias during model training and enables more stable regression-based learning.

In this study, we also introduce MahaSBERT-STS-v2\footnote{\url{https://huggingface.co/l3cube-pune/marathi-sentence-similarity-sbert-v2}}, a fine-tuned Sentence-BERT model for Marathi trained on the MahaSTS dataset. The base model, MahaSBERT, is trained on the IndicXNLI dataset alone. Prior work by \citet{joshi2023l3cube} has shown that sequential training, first on a Natural Language Inference (NLI) dataset followed by fine-tuning on an STS-style dataset, yields better performance than training on NLI alone.

We evaluate MahaSBERT-STS-v2 against several baselines, including MahaBERT, MuRIL, IndicBERT, and IndicSBERT, using Pearson and Spearman correlation coefficients to assess alignment with human judgments. Our results show that the MahaSTS dataset, combined with domain-specific fine-tuning, significantly improves performance on sentence similarity tasks for Marathi.

The main contributions of this work are as follows:

\begin{itemize}
    \item We introduce L3Cube-MahaSTS, the first human-annotated sentence similarity dataset based on real Marathi text, comprising 16,860 balanced sentence pairs.
    \item We release MahaSBERT-STS-v2, a fine-tuned Sentence-BERT model for Marathi, and show that task-specific training on MahaSTS significantly improves performance over existing baselines.
\end{itemize}

\section{Related work}

BERT (Bidirectional Encoder Representational Transfer) \citep{devlin2019bert}, a pre-trained transformer-based language model, has achieved state-of-the-art performance across various NLP tasks, including text classification and semantic textual similarity (STS).

\citet{reimers2019sentence} introduces Sentence-BERT (SBERT), a computationally efficient and fine-tuned BERT in a siamese network architecture. The authors show that training on NLI, followed by training on STSb leads to a marginal
improvement in performance.

Prior studies have demonstrated the effectiveness of monolingual language models over their multilingual counterparts when tailored for specific languages.  \citet{straka2021robeczech} presents a Czech monolingual RoBERTa model that significantly surpasses both multilingual models and other Czech-language models of similar size. \citet{scheible2020gottbert} shows that a German monolingual BERT model built on RoBERTa architecture outperforms various German and multilingual BERT models, even with minimal hyperparameter tuning.

Similarly, there have been a few studies for the Marathi language as well. In \citet{velankar2022mono}, researchers compare multilingual BERT-based models with their Marathi monolingual alternatives and report that the monolingual models deliver superior performance in single-language tasks by producing better sentence representations. \citet{joshi2022l3cube} introduces MahaFT—Marathi fastText embeddings trained on a monolingual corpus—which performs competitively against other publicly available fastText models. The SBERT models based on translated datasets for Marathi (MahaSBERT) and prominent Indic languages (IndicSBERT) were introduced in \citet{joshi2023l3cube} and \citet{deode2023l3cube} respectively. \citet{jadhav2025mahaparaphrase} further contributed by introducing the L3Cube-MahaParaphrase dataset, a high-quality human-annotated Marathi paraphrase corpus consisting of 8,000 sentence pairs, aiding semantic similarity and paraphrase identification tasks in low-resource settings. In this study, we use MahaSBERT as our base model, which is trained on the IndicXNLI\footnote{https://github.com/divyanshuaggarwal/IndicXNLI}  dataset, and then use it to finetune it further on the MahaSTS dataset, so that the performance of the base model can be enhanced to be used on STS-style datasets as well.

\citet{conneau2017supervised} demonstrated that supervised training on NLI tasks produces universal sentence representations that consistently outperform unsupervised baselines. \citet{dasgupta2018evaluating} proposed a new dataset targeting compositional-generalization in NLI. The authors discovered that augmenting training data with their compositional dataset improves performance on the new dataset without harming performance on existing benchmarks, highlighting the value of carefully structured datasets.

\citet{li2020sentence} analyzed the behavior of sentence embeddings extracted from pretrained models like BERT without fine-tuning. They proposed BERT-flow, a method that significantly improves performance across a variety of semantic textual similarity tasks, surpassing baseline sentence embedding methods. They also observed that native BERT similarity often correlates more with lexical overlap than true semantic similarity, but BERT-flow reduces that bias.

\citet{tang2018improving} proposes a shared multilingual sentence encoder pretrained using translation tasks, to perform semantic textual similarity tasks (STS) in low-resource languages by leveraging annotated data from high-resource languages to significantly outperform non-MT baselines on SemEval STS.

\citet{feng2020language} introduces LaBSE\footnote{https://huggingface.co/setu4993/LaBSE} (Language-agnostic BERT Sentence Embedding), a dual-encoder model pre-trained with MLM, TLM, trained on parallel multilingual data to produce language-agnostic sentence embeddings and enabling high-quality semantic similarity and retrieval for 109 languages.

\section{Dataset Curation}

MahaSTS is a human-annotated sentence similarity dataset designed to support fine-grained semantic similarity modeling in Marathi. It comprises 16,860 sentence pairs, each annotated with a continuous similarity score in the range of 0 to 5. To ensure balanced supervision and reduce label bias, the dataset is uniformly distributed across six predefined buckets. Each bucket contains exactly 2,810 sentence pairs and corresponds to a specific semantic similarity range, defined as follows:
\begin{itemize}
\item \textbf{Bucket 0 (score = 0):} The sentence pair shares no semantic similarity or is completely unrelated in meaning.
\item \textbf{Bucket 1 (0.1–1.0):} The pair has minimal similarity, possibly overlapping in a few words but differing significantly in meaning.
\item \textbf{Bucket 2 (1.1–2.0):} The sentences are somewhat related, with partial thematic or topical overlap but different in intent or details.
\item \textbf{Bucket 3 (2.1–3.0):} The pair exhibits moderate similarity, conveying related information with some variation in expression or focus.
\item \textbf{Bucket 4 (3.1–4.0):} The sentences are highly similar, differing only slightly in structure or specific content.
\item \textbf{Bucket 5 (4.1–5.0):} The pair is nearly or fully semantically equivalent, with only minor lexical or syntactic variations.
\end{itemize}

This structured bucketing enables the dataset to capture a wide spectrum of semantic relationships while maintaining an even distribution of training examples across similarity levels. Such design promotes more stable and generalizable regression performance during model training and evaluation.


\subsection{Data collection}
For the STS corpus creation task, we used the 1M real Marathi sentences from the L3Cube-MahaCorpus dataset \citet{joshi2022l3cube}. We preprocessed the 1M sentences, removing too short(< 3 words), too long (>20 words), non-Marathi language sentences, and duplicates. In this way, we were left with good sentences. We created embeddings of all sentences in the file using MahaSBERT-STS \footnote{https://huggingface.co/l3cube-pune/marathi-sentence-similarity-sbert}. We picked 5000 random sentences from the 1M corpus as query sentences. We then calculated the cosine similarity between each query sentence and the 1M corpus sentences.

\begin{figure*}[hbtp]
        \centering
        \includegraphics[scale=0.65]{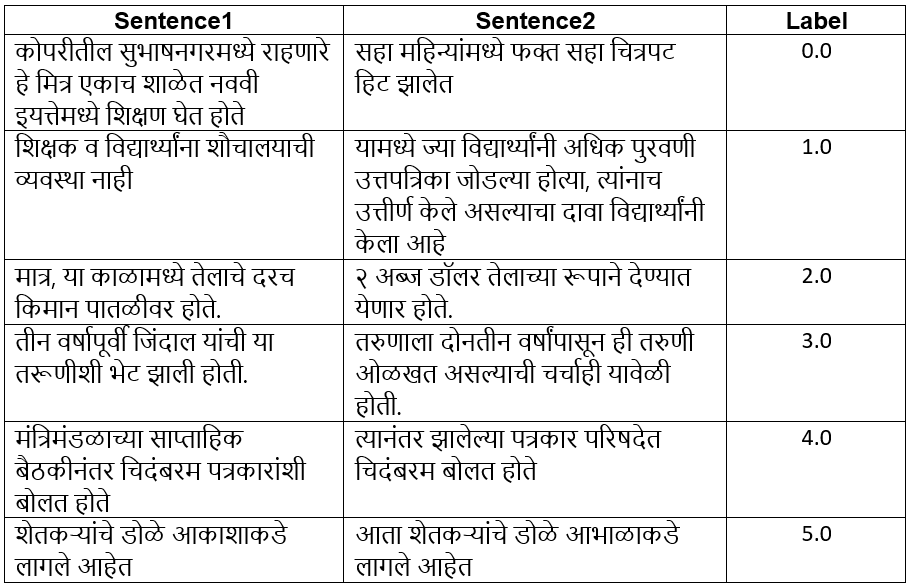}
        \caption{Examples of sentence pairs with labels in the range 0-5 from the L3Cube-MahaSTS dataset.}
        \label{fig:example_sentences}
    \end{figure*}

\subsection{Data preprocessing}
To preprocess the data, we divided the cosine scores in 5 buckets: (0.8, 1] went into bucket 5, (0.6, 0.8] went into bucket 4, (0.4, 0.6] went into bucket 3, (0.2, 0.4] went into bucket 2, (0, 0.2] went into bucket 1. We saved 1 similar sentence per bucket, for each of the 5,000 query sentences. In this way, we got 5,000 query sentences x 1 similar sentence x 5 buckets = 25,000 pairs. The sentence pairs that were completely dissimilar were eventually put into a 6th bucket, Bucket 0.

After we got the 25,000 pairs, we started annotating the sentences, using the cosine similarity scores from MahaSBERT-STS as a reference to help us with the annotations. During this process, we found some sentences to be incomplete or not making much sense, so we discarded those pairs. We were then left with 16,860 good sentence pairs. Figure \ref{fig:example_sentences} shows some examples of the sentence pairs from the MahaSTS dataset.

\subsection{Dataset statistics}
In the MahaSTS dataset, we have 16,860 pairs of sentences in total. There are 2,810 sentence pairs in each bucket, including bucket 0. The MahaSTS dataset was split into the train, test, and validation datasets in the ratio of 85:10:5. The train dataset contains 14,328 sentence pairs, test dataset contains 1,692 sentence pairs and the validation dataset has 840 sentence pairs. Each bucket in the train split contains 2,388 (0.85x) sentence pairs, in the test split contains 282 (0.1x) sentence pairs and in the validation split contains 140 (0.05x) sentence pairs where x = 2,810 which is the total number of sentence pairs in a single bucket in the MahaSTS dataset. Table \ref{tab:dataset_distribution1} presents the distribution of the data in train, test and validation datasets.

\begin{table}[hbtp]
    \centering
    \begin{tabular}{ccc}
    \hline
       \textbf{Dataset} & \textbf{Each bucket (0-5)}   & \textbf{Total}\\
    \hline
    {Train} & {2388} & {14328} \\
    {Test} & {282} & {1692} \\
    {Validation} & {140} & {840} \\
    \end{tabular}
    \caption{Number of sentence pairs in each bucket in the train, test and validation split and the total number of sentence pairs in the split.}
    \label{tab:dataset_distribution1}
\end{table}

\begin{table}[hbtp]
    \centering
    \begin{tabular}{cccc}
    \hline
       \textbf{Train}  & \textbf{Test} & \textbf{Validation} & \textbf{Total}\\
    \hline
    {14328} & {1692} & {840} & {16860}
    \end{tabular}
    \caption{Distribution of MahaSTS dataset into train, test, and validation splits in the ratio 85:10:5.}
    \label{tab:dataset_distribution2}
\end{table}

\begin{table*}[hbtp]
  \centering
  \begin{tabular}{lll}
    \hline
    \textbf{Model} & \textbf{Pearson coefficient} & \textbf{Spearman coefficient} \\
    \hline
    l3cube-pune/marathi-sentence-bert-nli & \textbf{0.9600} & \textbf{0.9523} \\
    l3cube-pune/marathi-bert-v2 & 0.9483 & 0.9386 \\
    google/muril-base-cased & 0.9361 & 0.9267 \\
    ai4bharat/indic-bert & 0.7311 & 0.7004 \\
    l3cube-pune/indic-sentence-bert-nli & 0.9515 & 0.9441 \\
  \end{tabular}
  \caption{Pearson correlation coefficient and Spearman correlation coefficient of different models trained on the MahaSTS dataset.}
  \label{tab:model_performance}
\end{table*}

\begin{table*}
  \centering
  \begin{tabular}{lll}
    \hline
\textbf{Pooling strategy} & \textbf{Pearson coefficient} & \textbf{Spearman coefficient} \\
\midrule
CLS & 0.958 & 0.9503 \\
MEAN & \textbf{0.9600} & \textbf{0.9523} \\
MAX & 0.9532 & 0.9444\\
\bottomrule
\end{tabular}
\caption{Performance of l3cube-pune/marathi-sentence-bert-nli model using different pooling strategies.}
\label{tab:pooling_strategies}
\end{table*}

\section{Models}

\subsection{MahaSBERT}

The MahaSBERT\footnote{https://huggingface.co/l3cube-pune/marathi-sentence-bert-nli} (l3cube-pune/marathi-sentence-bert-nli) is a Marathi sentence BERT model provided by L3Cube. It is a model trained on the IndicXNLI dataset using the MahaBERT model as the base model.

\subsection{MahaBERT}

MahaBERT\footnote{https://huggingface.co/l3cube-pune/marathi-bert-v2} (l3cube-pune/marathi-bert-v2) is a Marathi BERT model. It is a multilingual BERT (google/muril-base-cased) model fine-tuned on L3Cube-MahaCorpus and other publicly available Marathi monolingual datasets.\footnote{https://github.com/l3cube-pune/MarathiNLP}

\subsection{MuRIL}

Multilingual Representations for Indian Languages -- MuRIL\footnote{https://huggingface.co/google/muril-base-cased}(google/muril-base-cased) is a BERT model pre-trained on 17 Indian languages. This model uses a BERT base architecture.

\subsection{IndicBERT}

IndicBERT\footnote{https://huggingface.co/ai4bharat/indic-bert} (ai4bharat/indic-bert) is a multilingual ALBERT model pretrained on 12 Indian languages. This model is trained on the monolingual corpus provided by AI4Bharat.

\subsection{IndicSBERT}

IndicSBERT\footnote{https://huggingface.co/l3cube-pune/indic-sentence-bert-nli} (l3cube-pune/indic-sentence-bert-nli) is a sentence BERT model provided by L3Cube for 10 Indic languages. It is based on the MuRIL model, further fine-tuned on the NLI dataset of 10 major Indian languages.

\section{Results and Discussion}

We train and evaluate the models described in Section 3 on the MahaSTS dataset. Three different embedding strategies are explored: CLS embeddings, MEAN embeddings, and MAX embeddings. Among these, the MEAN pooling strategy consistently yields the best performance. The performance of the pooling strategies is presented in Table \ref{tab:pooling_strategies}.
Our primary model, MahaSBERT, is first fine-tuned on the training split of the MahaSTS dataset, which consists of 14,328 sentence pairs in the form (sentence1, sentence2, label). Training is performed using the CosineSimilarityLoss function, for 2 epochs, with a batch size of 8, AdamW optimizer, a learning rate of 1e-5, and the MEAN pooling strategy.

After training, the model is evaluated on the test set of the MahaSTS dataset containing 1,692 labeled Marathi sentence pairs. We use the Pearson and Spearman correlation coefficients as evaluation metrics. Our fine-tuned model, MahaSBERT-STS-v2, achieves a Pearson score of \textbf{0.9600} and a Spearman score of \textbf{0.9523}, representing a notable improvement over the original MahaSBERT model (Pearson: 0.9355, Spearman: 0.9268).

Table \ref{tab:model_performance} presents the results of all evaluated models (from Section 3) on the test set of our curated dataset. The MahaSBERT model outperforms all other models in terms of both Pearson and Spearman scores. Table \ref{tab:pooling_strategies} highlights the impact of different pooling strategies on MahaSBERT. As shown, the MEAN pooling approach outperforms CLS and MAX pooling strategies.
In conclusion, MahaSBERT, trained on the MahaSTS dataset using MEAN pooling, achieves the best results among all models evaluated in this study, confirming its effectiveness for sentence similarity tasks in Marathi.



\section{Conclusion}

In this work, we introduce the MahaSTS dataset, a human-annotated Sentence Textual Similarity (STS) dataset in the form (sentence1, sentence2, label). Each sentence pair is labeled with continuous similarity scores in the range of 0–5. The six score-based buckets are uniformly distributed across the train, test and validation splits to ensure there is minimal bias during training and evaluation. We also introduce the MahaSBERT-STS-v2 model, trained and fine-tuned on the MahaSTS dataset and evaluated using Pearson and Spearman correlation coefficients. The fine-tuned model outperforms the other models evaluated as a part of this study. Our results demonstrate the importance of human-annotated datasets and the results of structured supervision for low-resource languages.

\section*{Limitations}

There is limited generalization for longer or more complex sentences. Sentence-BERT models, especially in Indic contexts, tend to work best on short to moderately long sentences. For complex or compound sentences in Marathi, semantic representations may degrade. These limitations can be overcome by creating separate datasets containing sentences of varying lengths.

 \section*{Acknowledgement}
This work was carried out under the mentorship of L3Cube, Pune. We would like to express our gratitude towards our mentor, for his continuous support and encouragement. This work is a part of the L3Cube-MahaNLP project \citep{joshi2022l3cube}.

\bibliography{main}

\end{document}